\documentclass{article}

\usepackage{arxiv}

\usepackage[utf8]{inputenc} % allow utf-8 input
\usepackage[T1]{fontenc}    % use 8-bit T1 fonts
\usepackage{hyperref}       % hyperlinks
\usepackage{url}            % simple URL typesetting
\usepackage{booktabs}       % professional-quality tables
\usepackage{amsfonts}       % blackboard math symbols
\usepackage{nicefrac}       % compact symbols for 1/2, etc.
\usepackage{microtype}      % microtypography
\usepackage{lipsum}		% Can be removed after putting your text content
\usepackage{graphicx}
\usepackage{natbib}
\usepackage{doi}
\usepackage{tabularx}
\usepackage{placeins}
\usepackage{pdflscape}
\usepackage[justification=centering]{caption}
\usepackage{indentfirst}
\usepackage{afterpage}
\usepackage{amsmath,amssymb,amsfonts}
\usepackage{geometry}
\usepackage{multirow}
\usepackage{float}

\title{Deep semi-supervised approach based on
 consistency regularization and similarity learning
 for weeds classification}

\author{ \href{https://orcid.org/0009-0007-3053-1805}{\includegraphics[scale=0.06]{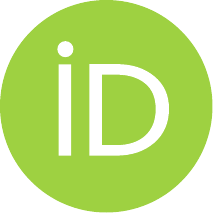}\hspace{1mm}Farouq Benchallal} \\
	PRISME EA 4229\\
	INSA CVL\\
	Bourges, 18002, Centre Val de Loire, France \\
	\texttt{farouq.benchallal@insa-cvl.fr} \\
	\And
	\href{https://orcid.org/0000-0003-3185-9996}{\includegraphics[scale=0.06]{orcid.pdf}\hspace{1mm}Adel Hafiane} \\
	PRISME EA 4229\\
	INSA CVL\\
	Bourges, 18002, Centre Val de Loire, France \\
	\texttt{adel.hafiane@insa-cvl.fr} \\
	\AND
	\href{https://orcid.org/0000-0003-2321-942X}{\includegraphics[scale=0.06]{orcid.pdf}\hspace{1mm}Nicolas Ragot}\\
	LIFAT EA 6300 \\
	Université Tours\\
	Tours, 37200, Centre Val de Loire, France \\
	\texttt{nicolas.ragot@univ-tours.fr} \\
	\And
	\href{https://orcid.org/0000-0001-9100-7539}{\includegraphics[scale=0.06]{orcid.pdf}\hspace{1mm}Raphael Canals}\\
	PRISME EA 4229 \\
	Université d’Orleans\\
	Orleans, 45067, Centre Val de Loire, France \\
	\texttt{raphael.canals@univ-orleans.fr} \\
}

\renewcommand{\shorttitle}{Deep semi-supervised approach based on
 consistency regularization and similarity learning}

\hypersetup{
pdftitle={Deep semi-supervised approach based on
 consistency regularization and similarity learning
 for weeds classification},
pdfsubject={Deep learning, Semi-supervised learning, Precision Agriculture},
pdfauthor={Farouq Benchallal, Adel Hafiane, Nicolas Ragot, Raphael Canals},
pdfkeywords={Semi-supervised learning, Deep learning, Consistency regularization, Similarity learning},
}

\begin{document}
\maketitle

\begin{abstract}

Weed species classification represents an important step for the development of automated targeting systems that allow the adoption of precision agriculture practices. To reduce costs and yield losses caused by their presence. The identification of weeds is a challenging problem due to their shared similarities with crop plants and the variability related to the differences in terms of their types. Along with the variations in relation to changes in field conditions. Moreover, to fully benefit from deep learning-based methods, large fully annotated datasets are needed. This requires time intensive and laborious process for data labeling, which represents a limitation in agricultural applications. Hence, for the aim of improving the utilization of the unlabeled data, regarding conditions of scarcity in terms of the labeled data available during the learning phase and provide robust and high classification performance. We propose a deep semi-supervised approach, that combines consistency regularization with similarity learning. Through our developed deep auto-encoder architecture, experiments realized on the DeepWeeds dataset and inference in noisy conditions demonstrated the effectiveness and robustness of our method in comparison to state-of-the-art fully supervised deep learning models. Furthermore, we carried out ablation studies for an extended analysis of our proposed joint learning strategy.

\end{abstract}

% keywords can be removed
\keywords{ Semi-supervised learning \and Deep learning \and Consistency regularization \and
 Similarity learning \and Precision Agriculture}

\section{Introduction}

Weeds result in the highest potential yield loss in crops, along with pathogens (fungi, bacteria, 
etc.) and animal pests (insects, rodents, nematodes, mites, birds, etc.) being both of lesser 
importance \citep{bib1}. The losses attributed to weeds are related to several factors such as, 
their time of emergence, their density and type \cite{bib2}. Weed species are highly 
competitive with crops for natural resources and characterized by a far better reproductive 
mechanism compared to crop plants. As a result, these unsown plant species constantly invade 
fields to overcome crops \cite{bib3,bib4}. The uncontrolled growth of 
weeds can lead to substantial economic losses \cite{bib2,bib5,bib6,bib7}. Therefore, addressing this problem is essential to meet the demands of an increasing global population. Currently, most weed management 
practices rely on the use of herbicides for the control of weeds. However, the over dependance  on the application of herbicides with similar modes of action led to the development of herbicide-resistant weeds \cite{bib8}. The growing concerns about the excessive use of agrochemicals \cite{bib9} incentivize the adoption of precision agriculture practices \cite{bib10}.

The development of automated systems for weed control requires accurate identification and recognition of weed species. This is a challenging problem due to field conditions, such as changes in lighting and illumination and some cases the resemblance between weeds and crop plants regarding their color, texture and morphological characteristics \citep{bib11}. Employing machine learning approaches for weed species classification involves an extensive process of feature extraction and selection followed by the application of machine learning classifiers. Sabzi et al.~\cite{bib12} extracted texture, color and shape features, along with five moment-invariant features from ground-based images. Subsequently, they applied different algorithms to select discriminative features, followed by weed classification using machine learning based classifiers, such as support vector machines and random forests. Machine learning methods require deep domain expertise to carry out time-intensive process of feature engineering. Conversely, deep learning approaches are distinguished by their strong ability to automatically extract discriminative features from data through representation learning \cite{bib13}. The high learning capacity of the deep learning models enables them to perform classification and prediction remarkably well, given enough labeled data. 

Currently, many successful deep neural network architectures \cite{bib14} have been used in agricultural applications. Reedha et al.~\cite{bib15} applied transformer-based networks, namely ViT-B-16 and ViT-B-32 for the classification of weeds and crop plants by means of unmanned aerial vehicles (UAVs). Ahmed et al.~\cite{bib16} utilized convolutional neural networks (CNNs) such as ResNet50 \cite{bib17} and InceptionV3 \cite{bib18} for the identification of weed species within the fields of soybean and corn from ground-based imagery. Valente et al.~\cite{bib19} applied seven variants of pre-trained CNN-models for the detection of weeds in grassland using UAV imagery. Based on cross validation method, the model MobileNet \cite{bib20} provided the best detection performance. Mesías-Ruiz et al.~\cite{bib21} utilized UAV images for weed species identification during the early growth stages in crop fields of maize and tomato, leveraging three types of CNNs i.e., VGG16 \cite{bib22}, ResNet152 \cite{bib17}  and Inception-ResNet-V2 \cite{bib23}. In relation to the different training scenarios, Inception-ResNet-V2 achieved the highest performance.  

The new developed saliency detection approaches \cite{bib24, bib25} favor deep learning-based techniques to discern salient objects and important regions in images. This is realized by relying on the multi-level features derived from deep neural networks to generate saliency maps.  Zhao et al.~\cite{bib26} proposed pyramid feature attention network to focus on high level context-aware features and on low-level spatial structural features, which were then both used to generate improved saliency maps. Furthermore, saliency detection can strengthen image classification and segmentation.  Zeng et al.~\cite{bib27} developed a network that is composed of a segmentation model and saliency aggregation module to capture the connections between the two computer vision tasks. As for co-saliency detection, the goal is to detect salient regions from a set of related images by exploring the interdependencies between the images. Wei et al.~\cite{bib28} presented a deep learning technique based on a fully convolutional network to detect co-salient objects. Similarly, co-segmentation aims to segment common objects from a group of relevant images. Li et al.~\cite{bib29} proposed a deep co-segmentation method to segment common objects belonging to a common semantic class from image pairs.

Nevertheless, to exploit the full potential of the deep learning algorithms, large, labeled datasets are needed, this represents a limitation for the agricultural applications as a consequence of the difficulties related to the annotation process. On the other hand, unlabeled data could be acquired in copious amounts with relative simplicity. The challenges to overcome regarding semi-supervised learning (SSL) is to perform similarly or surpass the state-of-the-art deep learning supervised models on datasets that are highly constrained in terms of labeled training data. A specific way to attain this is by solving the dilemma between leveraging an accurate model with enough parameters, considering that only a small amount of labeled data will be available for its training. 

Our proposition is aligned with this direction. Henceforth, in this paper we propose a method that employs SSL to improve the way of utilizing the unlabeled data, in conditions where the labeled data is exceedingly more restricted. Our main contributions are outlined as follows: 

	\begin{itemize}
 
		\item 	A deep semi-supervised method combining consistency regularization and similarity learning for effective and robust weed species recognition. 
		\item 	An auto-encoder architecture, based on a ConvNeXt Encoder and specifically designed de
coder with skip-connections. Through our joint learning strategy, it allows an increased ability
 for incorporating the most relevant information from unlabeled data in scenarios where the
 labeled data are excessively scarce. 
		\item 	Extensive experiments on the public dataset DeepWeeds for a rigorous assessment and evaluation of our proposed approach to demonstrate its effectiveness and behavior. 
	\end{itemize}

The remainder of this paper is organized as follows. In the next section we present the related works. The subsequent section describes comprehensively our proposed method. The fourth section details at length the carried-out experiments alongside an in-depth analysis of the obtained results. Finally, our conclusions and perspectives are summarized in the last section. 

\section{Related Works}\label{sec2}

In the literature there are studies, which adopt the semi-supervised paradigm for the elaboration of methods with the objective of identifying and recognizing weed species. Kerdegari et al.~\cite{bib30} introduced a semi-supervised generative adversarial network (GAN) for the semantic segmentation of weeds and crops using aerial multispectral imagery. Khan et al.~\cite{bib31} devised an optimized semi-supervised GAN-based framework for the identification of weeds and crops utilizing UAV images. Shorewala et al.~\cite{bib32} applied a semi-supervised method which encompasses two steps. An unsupervised binary segmentation step to generate vegetation masks, followed by a classification step for distinguishing weeds and crops by means of a finetuned CNN model from ground-based imagery.  Homan and du Preez.~\cite{bib33} proposed a two fold method that is comprised of feature recognition along with the classification of plant species through deep semi-supervised learning. Hu et al.~\cite{bib34} combined image synthesis with semi-supervised learning for the training of site-specific models to enable the detection of weeds. Liu et al.~\cite{bib35} developed a semi-supervised approach that incorporates a mechanism of mixed attention to enhance the model’s ability in extracting important features to detect weed species. Kong et al.~\cite{bib36} introduced the semi-supervised learning paradigm for the detection of weeds in wheat, showing that a full supervised method needs sufficient labeled data to achieve satisfactory performance compared to the SSL-based method. Similarly, Chen et al.~\cite{bib37} used an SSL method for the detection of weeds in sod farms. Deng et al.~\cite{bib38} proposed a semi-supervised object detection method called WeedTeacher to benefit from the unlabeled data for enhanced detection of weeds in-domain and cross-domain contexts. 

As aforementioned we developed our semi-supervised approach incorporating consistency regularization with similarity learning. With respect to the incorporation of consistency regularization we were mainly influenced by the ladder network \cite{bib39} due to its successful utilization of consistency regularization. Subsequently, several other consistency regularization methods were introduced, such as the Pi-model and temporal ensembling \cite{bib40}. The Pi-model relies on stochastic transformations applied to the training samples for the integration of consistency regularization. Temporal ensembling relies on both applying stochastic transformations as well as leveraging an exponential moving average of the predictions. Following these methods the mean-teacher \cite{bib41} was proposed, which relies more on the structure of the networks for the integration of consistency regularization. We should note that the methods that are based on consistency regularization suffer from confirmation bias \cite{bib42} due to the dependance on a single model within the semi-supervised architecture to generate the predictions for consistency training. In the case where these predictions (corresponding to the unlabeled training samples) are incorrect across multiple iterations, this would lead to a negative impact on the semi-supervised learning performance. To mitigate the effect of confirmation bias, Ke et al.~\cite{bib43} presented an approach called dual student, which utilizes a stability constraint with respect to the unlabeled data to address the limitation of confirmation bias.

Relative to similarity learning we were motivated by methods that leverage joint learning for enhancing the predictive ability of the learning models. For instance, the deep learning framework introduced by \cite{bib44}, considers both class label information and local spatial distribution information between training samples through a pairwise loss to constrain representation learning. Similarly, Paclik et al.~\cite{bib45} proposed an approach which employs a trainable similarity measure to build second-stage classifiers for object detection. The measure is based on local matches in a set of regions within an image to increase robustness. Horn and Muller \cite{bib46} proposed a neural network architecture termed similarity encoder constructed for simultaneously learning a mapping from an original input feature space into similarity preserving embedding space, while factorizing a target matrix with pairwise relations.  Xia et al.~\cite{bib47} developed a method that utilizes the cosine similarity metric for similarity learning, which is based on an ensemble of cosine learners.

Regarding our approach, we relied more on the design of our proposed deep encoder-decoder architecture and the utilization of skip-connections for the integration of consistency regularization for the purpose of limiting confirmation bias by providing reliable information during the training from the labeled and unlabeled data. Moreover, to leverage our proposed architecture and develop our learning strategy, we were interested in how to further improve the consistency regularization training in relation to the scarcity of the labeled training data, by focusing on the incorporation of similarity learning into the training of the deep semi-supervised models to guide the process of joint learning toward improving the generalization ability of the trained deep semi-supervised models. Finally, the consistency-based methods mentioned above were applied to general purpose datasets. In contrast, for the development of our approach we were also motivated by the effective recognition and identification of weed species from images acquired in real world conditions.

\begin{landscape}
\begin{figure}
	\centering
	\captionsetup{justification=centering}
	\includegraphics[width=21.4cm,height=11.6cm]{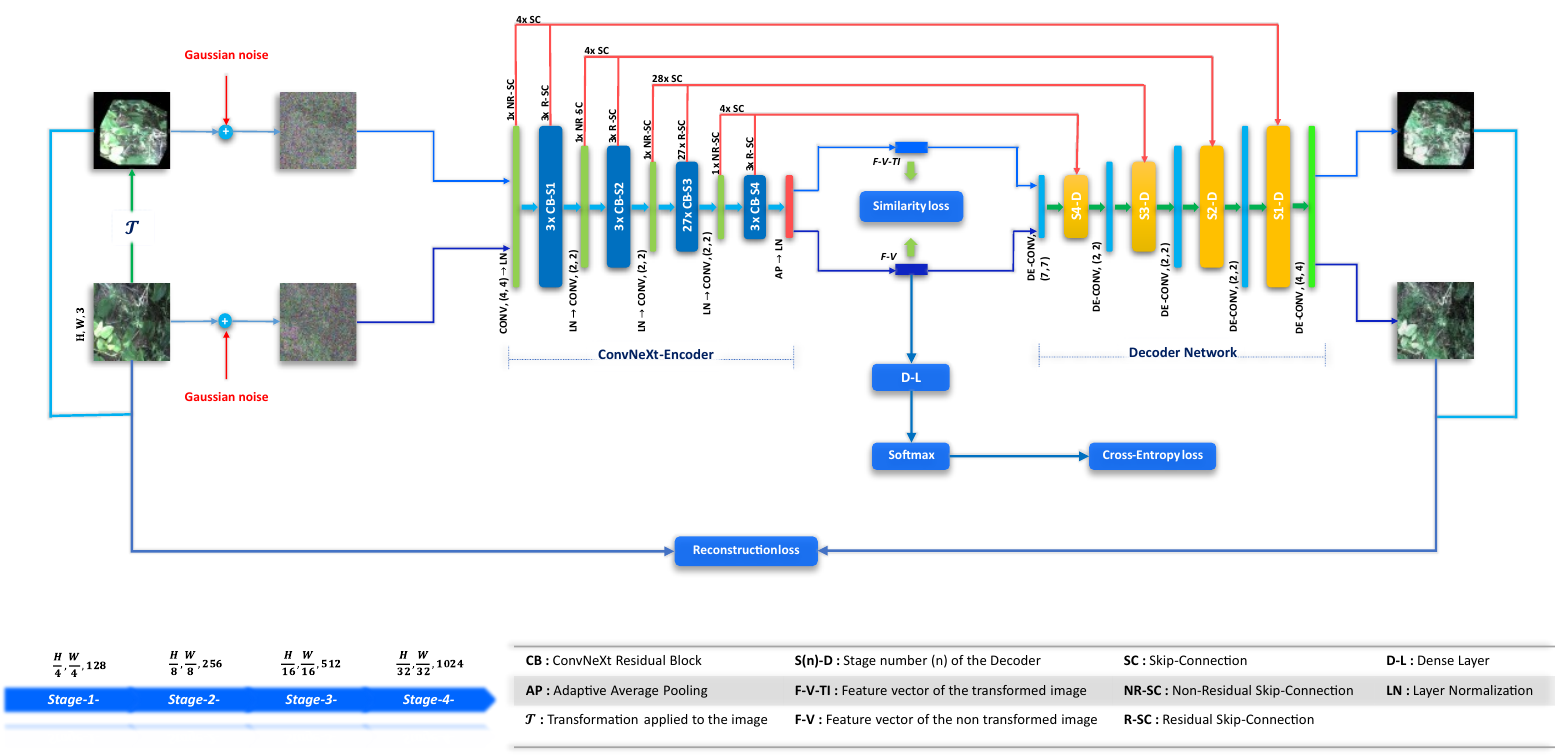}
	\caption{Overview of the semi-supervised approach}
	\label{Figure 1}
\end{figure}
\end{landscape}

\section{Proposed method}\label{sec3}

\subsection{Problem statement}\label{sec3.1}

Supervised learning involves a sequence of training pairs $\left( \mathbf{x_{1}}, y_{1} \right), \left( \mathbf{x_{2}}, y_{2} \right), \ldots , \left( \mathbf{x_{n}}, y_{n}\right)$, where each pair $\left( \mathbf{x_{i}}, y_{i} \right)$ is composed of an input $\mathbf{x_{i}}\in\mathcal{X}$ from the space of inputs, and a label $y_{i}$ $\in$ $\mathcal{Y}$ from the space of outputs. In our case $\mathcal{X}\subset\mathbb{R}^{d}$ represents the set of images, and we have $\mathcal{Y}=\left\{1,\ldots,C\right\}$ representing the species of weeds that we want to identify. C refers to the number of classes. The joint space $\mathcal{X}\times\mathcal{Y}$ is assumed to be a probabilistic space with an unknown probability measure $\mathcal{P}\left(\mathbf{x}, y\right)$ and the data is sampled from this space. The joint measure $\mathcal{P}\left(\mathbf{x}, y\right)$ can be decomposed into a measure of the marginal $\mathcal{P}\left(\mathbf{x}\right)$, and a measure of the conditional distribution $\mathcal{P}\left(y\middle|\mathbf{x}\right)$. Supervised learning aims at estimating a functional relationship $\mathbf{x} \longrightarrow y$ between a covariate $\mathbf{x}\in\mathcal{X}$ and the class variable $y\in\left\{1,\ldots,C\right\}$, with the goal of minimizing the classification error. In semi-supervised learning settings, in addition to the labeled data $\mathcal{D}_\ell=\left\{\left(\mathbf{x_{i}},y_{i}\right)\middle|i=1,\ldots, n\right\}$ sampled from $\mathcal{P}\left(\mathbf{x}, y\right)$, we have also access to the unlabeled data $\mathcal{D}_{u}=\left\{\mathbf{x_{n+j}}\middle| j=1,\ldots, m \right\}$ sampled from $\mathcal{P}\left(\mathbf{x}\right)$. While acquiring labeled samples is expensive and time-intensive, gathering unlabeled samples can be done at low-cost and quickly. Then, $n\ll m$ which means that the size of the labeled portion could be much smaller than the size of the unlabeled portion \cite{bib48}. The objective of semi-supervised learning is to leverage the labeled data $\mathcal{D}_\ell$, with additional information about the data distribution $\mathcal{P}\left(\mathbf{x}\right)$, giving from $\mathcal{D}_{u}$ with the purpose of increasing the performance and achieving better generalization to new unseen samples \cite{bib49,bib42}. Semi-supervised learning requires that the data distribution should be under a set of assumptions \cite{bib48,bib42}, otherwise, the prediction's performance may not be improved. The main assumptions associated with semi-supervised learning are the following:

Semi-supervised smoothness assumption: if two points $\mathbf{x_{1}},\mathbf{x_{2}}$ in a high-density region are close, then so should be the corresponding outputs $y_{1},y_{2}$.  

Cluster assumption: If points are in the same cluster, they are likely to be of the same class, which means that the decision boundary should lie in a low-density region. 

Manifold assumption: high-dimensional data lie approximately on a low-dimensional manifold. In high-dimensional spaces volume grows exponentially with the number of dimensions, which makes it difficult to estimate reliable densities. If the data lie on a low dimensional manifold, then the learning algorithms can overcome the problems related to high dimensionality, by operating in the corresponding low-dimensional space. 

\clearpage

\subsection{Architecture of the proposed Auto-Encoder}\label{sec3.2}

In the current section we describe in-depth both the encoder and decoder models utilized in our proposed deep auto-encoder architecture, depicted in Figure \ref{Figure 1}. This is followed by a detailed explanation of our semi-supervised learning strategy. Finally, we provide more details regarding our method of integrating the similarity loss, for the joint optimization process.

\subsubsection{ConvNeXt-Encoder}\label{sec3.2.1}

ConvNeXt represents a family of pure convolutional neural networks devised through the gradual process of modernizing the standard family of RESNETS \cite{bib17} toward the design of vision transformers, through an in-depth exploration of the design spaces and the possible reachable limits, by relying solely on the use of CONVNET modules. The modernization process groups a series of design decisions applied on the macro/micro level-architecture \citep{bib50}. These series of modernization steps can be outlined in the following manner. The first step involves the distribution of computation throughout the network, meaning the adoption of a stage compute ratio that is similar to the design of vision transformers \cite{bib51}, which specifies the number of blocks that will be attributed to each stage. Within a multi-stage design, the feature maps resolution is progressively changed across stages. 
The second step is related to the stem cell design, which is used for processing the input images at the start of the network. Following the design of hierarchical vision transformers, the RESNET style-stem cell is replaced with a patchify stem implemented through a $4\times4$ non-overlapping convolution layer.
In the third step a special case of grouped convolution is adopted where the number of groups is equal to the number of channels. Depth-wise convolution \cite{bib20,bib52,bib53} bears resemblance to the weighted sum operation in self-attention, which mixes information only along the spatial dimension. Combining depth-wise convolution and $1\times1$ convolution results in a separation between the mixing of information along the spatial and the channel dimensions, which is an important property of vision transformers.
Within the fourth step, the design of inverted bottleneck with an expansion rate equal to 4 is used. The decision was motivated by the design of the transformer block \citep{bib51, bib54} where an inverted bottleneck is created due to the hidden dimension of the MLP (MultiLayer Perceptron) block which is four times wider than the input dimension.
In the fifth step, to benefit from large size kernels, the depth-wise convolution layer was positioned at the beginning of the residual block, followed with $1\times1$ convolution layer. This design choice is also inspired by the vision transformer block, in which we observe that the MSA (Multi-Head Self-Attention) module is positioned before the MLP layers. The size of the kernels associated with the depth-wise convolutional layer was increased from $3\times3$ to $7\times7$. 
Regarding the sixth step, the Gaussian Error Linear Unit (GELU) \cite{bib55}, which is the activation function utilized in most advanced transformers is used to replace the activation function Rectified Linear Unit (ReLU) \cite{bib56}. Subsequently, in order to replicate the style of the transformer block, all GELU activation functions within the residual block were removed with the exception of a single GELU activation function placed between two $1\times1$ convolution layers. Likewise, the number of batch normalization (BN) layers \cite{bib57} was also reduced to a single layer per block. Moreover, the batch normalization layer was substituted with Layer normalization (LN) \cite{bib58} following the normalization method applied in vision transformers.
As for the last step, to conduct the spatial downsampling between stages, separate downsampling layers were utilized instead of relying on the residual block at the start of each stage. And before each downsampling layer a LN layer was added. Regarding the patchify stem the normalization layer is added afterwards. The models of the ConvNeXt family achieve high-level performance across different model capacities (different types/variants of models) showing the effectiveness of the architecture design. The variant model that we have chosen for the encoder part of our proposed auto-encoder is called ConvNeXt-Base. We should note that the number of channels (the number of filters on the level of convolution layers) increases along the depth of the encoder as we move throughout the stages of the network.     

\subsubsection{Decoder network}\label{sec3.2.2}

Regarding the decoder network, we have adopted a multi-stage design with the aim of utilizing the feature maps with varying resolutions, from the different stages of the ConvNeXt encoder, for the reconstruction process. The feature maps on the encoder side are retrieved through residual and non-residual skip-connections. The residual skip-connections provide the decoder with direct access to the outputs of the residual blocks at any given stage of the encoder. The non-residual skip-connections enable the decoder to directly access the outputs of the downsampling layers including the patchify stem. The total number of skip-connections integrated into our architecture amounts to 40 (36 are residual skip-connections, corresponding to the ConvNeXt residual blocks and 4 non-residual skip-connections, corresponding to the downsampling layers and the patchify stem). We choose to utilize these two types of skip-connections to extract detailed information from the encoder path, which allows us to integrate the intermediate hidden representations (the encoder feature maps) in a manner that accounts for symmetry between the encoder and decoder, along with the dimensional change within the decoder path during the reconstruction process.
Consistently with the design of ConvNeXt-Base encoder, we have employed separate up sampling layers at the beginning of each decoder stage. These layers are essential for the structure of the decoder and for the process of integrating information from the skip-connections at each stage. Apart from the up-sampling layers, each decoder stage includes a specific number of de-convolution blocks. The number of these blocks corresponds to the number of skip-connections that connect the stage to the encoder path. Each de-convolution block receives as inputs the encoder representations from its respective skip-connection and the output of the preceding de-convolution block or the separate up-sampling layer. The source of the second input is determined by the position of the de-convolution block within the decoder stage. The first and the second inputs are then merged through element-wise summation, resulting in a mixing of the spatial information from both the decoder and the encoder. Subsequent to this step, a $1\times1$ de-convolution layer is applied with a stride of 1 and a number of kernels equal to the number of channels of the summed inputs. Due to the element-wise summation, the dimensions of the representations do not change within the de-convolution block. Thereafter, the output of the $1\times1$ de-convolution layer is normalized through layer normalization, and an activation function is applied in the final step to achieve the overall output of the block. With regard to the activation functions applied element-wise on the side of the decoder, we adopted LeakyReLU \cite{bib59} for stages 4, 3 and 2. For stage 1, we applied the ELU activation function \cite{bib60} that has improved learning characteristics and contributes to alleviate the vanishing gradient problem. As concluding element of the decoder, we added a final $4\times4$ de-convolution layer with a stride of 4 and 3 Kernels. An element-wise sigmoid activation function was then applied to generate the final output image. As part of preprocessing the pixel values of the input images were scaled between 0 and 1. Through dividing the channels of the images by 255. For semi-supervised learning settings a dense layer with a number of units equal to the classes that need to be recongnized is added. Which is applied to the feature vector obtained by leveraging the encoder model.

\subsection{Semi-supervised learning strategy}\label{sec3.3}

The semi-supervised optimization process is performed by minimizing the combined losses of both forms of data labeled and unlabeled. These combined losses comprise the following three distinct loss functions: supervised loss, consistency regularization loss (reconstruction from noisy inputs) and similarity loss. Their joint minimization using the available training data facilitates a refined guided optimization of the auto-encoder’s weights. The supervised loss is applied solely to the labeled portion of the training data, meaning that it relies on the supervised information provided by class labels of the annotated samples. Its minimization optimizes both the encoder model’s weights $\left(\theta_{E}\right)$ and the densely connected layer $\left(\theta_{d-l}\right)$ dedicated to classification. The cross-entropy function, defined below, is used to represent the supervised loss:

\begin{equation}
\mathnormal{l}_{\mathnormal{CE}}\left(\mathnormal{y}_\mathnormal{i},\mathnormal{f}\left(\mathbf{x_i}+\mathnormal{\zeta}_\mathnormal{i};\mathnormal{\theta}_\mathnormal{E},\mathnormal{\theta}_{\mathnormal{d}-\mathnormal{l}}\right)\right)=-\sum_{\mathnormal{j}=\mathnormal{0}}^{\mathnormal{C}-\mathnormal{1}}{\mathnormal{y}_{\mathnormal{ij}}\mathnormal{log}{\left(\mathnormal{f}_\mathnormal{j}\left(\mathbf{x_i}+\mathnormal{\zeta}_\mathnormal{i};\mathnormal{\theta}_\mathnormal{E},\mathnormal{\theta}_{\mathnormal{d}-\mathnormal{l}}\right)\right)}}
\end{equation}

\hfill \break

Here, $\mathnormal{y}_\mathnormal{i}$ denotes the ground truth label. $\mathnormal{f}\left(\mathbf{x_i}+\mathnormal{\zeta}_\mathnormal{i};\mathnormal{\theta}_\mathnormal{E},\mathnormal{\theta}_{\mathnormal{d}-\mathnormal{l}}\right)$ represents the predicted label $\widetilde{y}_{i}$ from the densely connected layer that employs the latent space from the encoder. $\mathnormal{\zeta}_\mathnormal{i}$ denotes the additive gaussian noise and C is the number of classes.  

The consistency regularization loss utilizes both types of data labeled and unlabeled. As we mentioned previously, through the minimization of this loss we seek to exploit the characteristics of our proposed architecture that enables a significant flow of information during the backward pass. To compute this loss, we have used the L2 loss function and its formulation taking into consideration both the non-transformed view of the input image $\left(\mathbf{x_i}\right)$ and its transformed view $\left(\mathbf{x_i}{'}\right)$ in the following way:

\begin{equation}
\mathnormal{l}_{\mathnormal{CR}}\left(\mathbf{x_i},\widetilde{\mathbf{x}}_{\mathbf{i}},\mathbf{x_i}{'},{\widetilde{\mathbf{x}}}_\mathbf{i}{'}\right)=(1/2)(\|\mathbf{x_i} - \widetilde{\mathbf{x}}_{\mathbf{i}}\|_{2}^{2}+\|\mathbf{x}_{\mathbf{i}}{'} - \widetilde{\mathbf{x}}_{\mathbf{i}}{'}\|_{2}^{2})
\end{equation}
\hfill \break

$\widetilde{\mathbf{x}}_{\mathbf{i}}$ denotes the reconstructed image of the non-transformed view and $\widetilde{\mathbf{x}}_{\mathbf{i}}{'}$ refers to the output reconstruction of the transformed view. 

As for the similarity loss, both types of data are utilized too. Regarding the measure of similarity between the high-level representations, obtained through the encoder model, we used a loss which relies on the cosine similarity function (\textbf{cos}) defined as follows:  

\begin{gather}
cos\left(\mathnormal{f}\left(\mathbf{x}_\mathbf{i}+\mathnormal{\zeta}_\mathnormal{i};\mathnormal{\theta}_\mathnormal{E}\right),\mathnormal{f}\left(\mathbf{x}_\mathbf{i}{'}+\mathnormal{\zeta}_\mathnormal{i}{'};\mathnormal{\theta}_\mathnormal{E}\right)\right)\nonumber\\
=\mathnormal{f}\left(\mathbf{x}_\mathbf{i}+\mathnormal{\zeta}_\mathnormal{i};\mathnormal{\theta}_\mathnormal{E}\right)\cdot\mathnormal{f}\left(\mathbf{x}_\mathbf{i}{'}+\mathnormal{\zeta}_\mathnormal{i}{'};\mathnormal{\theta}_\mathnormal{E}\right)/\max{\left(\|\mathnormal{f}\left(\mathbf{x}_\mathbf{i}+\mathnormal{\zeta}_\mathnormal{i};\mathnormal{\theta}_\mathnormal{E}\right)\|\cdot\|\mathnormal{f}\left(\mathbf{x}_\mathbf{i}{'}+\mathnormal{\zeta}_\mathnormal{i}{'};\mathnormal{\theta}_\mathnormal{E}\right)\|,\epsilon\right)}
\end{gather}

\hfill \break

With $\mathnormal{f}\left(\mathbf{x}_\mathbf{i}+\mathnormal{\zeta}_\mathnormal{i};\mathnormal{\theta}_\mathnormal{E}\right)$ and $\mathnormal{f}\left(\mathbf{x}_\mathbf{i}{'}+\mathnormal{\zeta}_\mathnormal{i}{'};\mathnormal{\theta}_\mathnormal{E}\right)$ representing the feature vectors (high-level representations) corresponding to the non-transformed input image and its transformed version, respectively. $\epsilon$ is a constant set to $10^{-8}$ for avoiding division by zero. Henceforth, the similarity 
loss is expressed in the following way: 

\begin{equation}
\mathnormal{l}_{\mathnormal{Sim}}\left(\mathnormal{f}\left(\mathbf{x}_\mathbf{i}+\mathnormal{\zeta}_\mathnormal{i};\mathnormal{\theta}_\mathnormal{E}\right),\mathnormal{f}\left(\mathbf{x}_\mathnormal{i}{'}+\mathnormal{\zeta}_\mathnormal{i}{'};\mathnormal{\theta}_\mathnormal{E}\right)\right)= 1-cos\left(\mathnormal{f}\left(\mathbf{x}_\mathbf{i}+\mathnormal{\zeta}_\mathnormal{i};\mathnormal{\theta}_\mathnormal{E}\right),\mathnormal{f}\left(\mathbf{x}_\mathbf{i}{'}+\mathnormal{\zeta}_\mathnormal{i}{'};\mathnormal{\theta}_\mathnormal{E}\right)\right)
\end{equation}

\hfill \break

Finally, the expression of the overall loss composed of the three loss functions is given in the following manner: 

\begin{equation}
\mathnormal{l}_{\mathnormal{total}}=\mathnormal{l}_{\mathnormal{CE}}+\lambda_{CR}\cdot\mathnormal{l}_{\mathnormal{CR}}+\lambda_{Sim}\cdot\mathnormal{l}_{\mathnormal{Sim}}
\end{equation}

\hfill \break

Where $\lambda_{CR}$ and $\lambda_{Sim}$ represent the two hyperparameters $\in[0,1]$ that determine the relative contribution of the two losses to the overall loss. 

\subsection{Reconstruction from noisy images and similarity learning }\label{sec3.4}
As mentioned previously we have opted to combine the process of reconstructing clean outputs from noisy inputs using the auto-encoder, through consistency regularization constraint with similarity learning to increase further the ability to exploit unlabeled data.  To enable this, the level where to incorporate the similarity term is important for the overall semi-supervised learning process. 

\begin{figure}[H]

	\centering
	\captionsetup{justification=centering}
	\includegraphics[width=13cm,height=2cm]{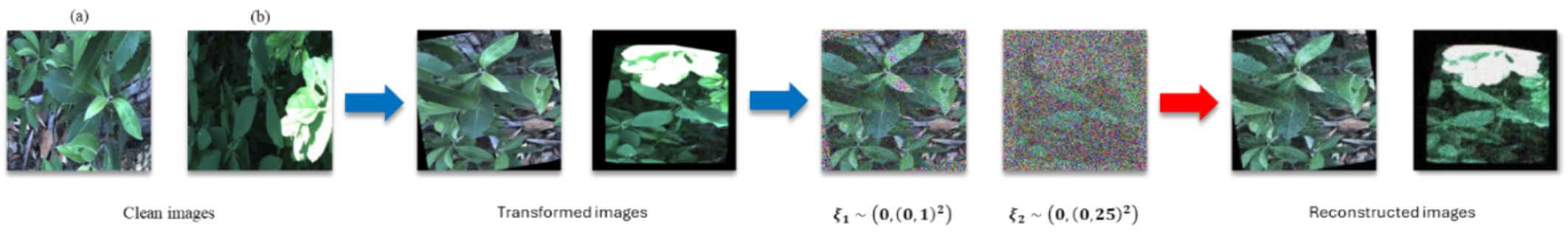}
	\caption{Resulting images reconstructed from noisy inputs through our proposed deep encoder-decoder model with similarity learning}

	\label{Figure 2}

\end{figure}

Hence, our choice, with respect to where we want to employ this term was motivated by our  interest in the extracted high-level abstract representations through the training data. These  high-level representations of original image and transformed image are distinct because they contain high-level context aware information that are rich semantically. Thereby, we have chosen to integrate the similarity term at the end of the encoder network to guide the optimization process toward learning the most important characteristics shared between high-level representations. Furthermore, one should note that the transformations applied to generate the transformed views of the training samples do have an influence on the learning process. For this reason, these transformations need to be selected thoroughly to help in acquiring beneficial information for the process of similarity optimization (see Figure~\ref{Figure 2}).

\section{Experiments and Results}\label{sec4}

This section introduces the experimental protocol and key evaluation metrics used for a thorough assessment of the proposed semi-supervised approach for weed species classification. The performance of our method is compared to state-of-the-art deep learning models that are characterized by having similar scale of trainable parameters, to highlight the benefits of the semi-supervised paradigm. 

\subsection{DeepWeeds dataset}\label{sec4.1}

For the purpose of training and evaluating the learning algorithms we have used the DeepWeeds dataset \cite{bib61} , which is constructed from images of weed species. Data were collected in-situ from eight different locations in northern Australia over a period spanning from  June 2017 to March 2018. Eight weed species were selected (target classes) that have notoriety for their invasiveness and their damaging impact. In order to reflect scene and target variability, several factors of variations needed to be taken into account during the collection of these images:  illumination, that varies throughout the day depending on the change in sunlight, which creates highly dynamic scenes; variability caused by complex and dynamic target backgrounds.

Furthermore, the dataset considers seasonal variation in the target weed species, which implies that a single class of weed species comprises images of the weed with and without flowers and in varying health conditions. These realistic environmental conditions constitute a significant challenge in the accurate classification of weed species. Additionally, DeepWeeds includes images of plant life native to the eight locations that represent negative class.  
Seeking to ensure the dataset diversity and breadth, more than 1,000 images per target weed species were gathered, with the objective of maintaining an approximate equal division of positive to negative class images across each location. In the end a total of 8,403 positive images of weed species and 9,106 negative images of neighboring flora and non-target backgrounds were collected. All the images within the DeepWeeds dataset were labeled by experts for the presence or absence of target weed species. 

\subsection{Deep learning models for comparative analysis}\label{sec4.2}

The supervised model \textbf{ConvNeXt-Base} \cite{bib50} is used as a basis for our encoder and is also used as a competitive approach. We have also incorporated two additional model variations from two separate neural network families: \textbf{EfficientNet-V2-L} \cite{bib62} and \textbf{ViT-B-16} \cite{bib51}. These models were selected based on their performances and the availability of pre-trained weights on the extensive \textbf{ImageNet-22K} dataset, a superset of \textbf{ImageNet-1K} comprising 14,197,122 images categorized into 21,841$\approx$ \textbf{22K} classes. Leveraging these pre-trained weights as a foundation for training on our target dataset offers considerable advantages, including improved performance and reduced training time compared to random  weight initialization.   
\textbf{EfficientNet-V2-L} belongs to the \textbf{EfficientNet-V2} family, based on combining the training-aware 
neural architecture search and scaling for an optimized training speed and parameter efficiency. 
\textbf{ViT-B-16} on the other hand is part of the vision transformers family.

\subsection{Settings for training and evaluation}\label{sec4.3}

\subsubsection{Hyperparameter setup and implementation details}\label{sec4.3.1}

With respect to the implementation of the deep learning algorithms and the running of the experiments, we have used the machine learning framework PyTorch \textbf{1.10.2} as well as the libraries PyTorch-Ignite \textbf{0.4.8} and timm \textbf{0.5.4} As for hardware specifications, we utilized the computing cluster CaSciModOT \cite{bib63}, which provided us access to three Nvidia Tesla V100 32 GB GPUs along with the AMD 7302 CPU. 

To ensure a fair comparison of the results obtained by the deep learning models, we aimed to use consistent optimal hyperparameters across all experiments. Therefore, to update model parameters, we have used the stochastic gradient descent algorithm (\textbf{SGD}) with momentum and damping set to zero. As for learning rate, we have used two different values \textbf{i.e.,} 0.01 and 0.009. The learning rate is then decreased by a factor of 0.9 after each 10 epochs. Regarding the utilization of pre-trained weights, we have replaced the dense layer of ImageNet-22K with a layer that has a number of units equal to the number of classes of DeepWeeds (\textbf{9} units). 
In relation to our semi-supervised training approach, we have used a gaussian noise with a standard deviation $\boldsymbol{\sigma}$ to 0.1 and a mean $\boldsymbol{\mu}$ set to zero. We have utilized pre-trained weights on the dataset ImageNet-1k, to to the initialize the decoders. We used two values for the $\boldsymbol{\lambda_{Sim}}$ hyperparameter in our loss function: 0.90 and 0.85. The $\boldsymbol{\lambda_{CR}}$ hyperparameter was set to 1.00. To generate more unlabeled data, we have also applied rotations varying from $-180^{\circ}$  to $+180^{\circ}$ to the training subset. The obtained images were then de-labeled during the semi-supervised training. 

\subsubsection{Similarity transformations}\label{sec4.3.2}
As we noted previously, the transformations employed in the context of similarity learning are selected with attention to bring a wider range of diversity to the generated views from both labeled and unlabeled training images. The transformations comes with a series of geometric and intensity transformations applied throughout the course of the training epochs. Geometric transformations encompass rotations in the range of $-120$ and $+120$ degrees. A random shift, both horizontally (by a factor of 0.2) and vertically (by a factor of 0.3) with a probability of 0.7 is also applied. Additionally, a random scaling ranging from 0.8 to 0.9 with a probability of 0.7 was performed. Finally, random horizontal and vertical flips were applied, each with a probability of 0.5. Intensity transformations consists in saturation transformations, which affect the intensity of colors, together with brightness transformations that change the illumination intensity. For the random sampling of the saturation factor, we have used two different values to represent the lower limit (min) of the sampling interval. The upper limit (max) was set to 2.6. In a similar manner we used the same low limit values for the random sampling of the brightness factor and the value 2.8 for the upper limit. The attributed probabilities for the saturation and brightness transformations are 0.6 and 0.7, respectively. 

\subsubsection{Evaluation procedure}\label{sec4.3.3}
During our experiments, we have employed the 5-fold stratified cross-validation method \cite{bib64,bib65,bib66}. Its aim is to determine the model’s ability to provide accurate predictions on new unobserved data samples. Stratified sampling was performed to ensure that each fold approximately reflects the original dataset’s class distribution. The initial partitioning of the dataset was done as follows: \textbf{60\%} were allocated for training, \textbf{20\%} used for validation and the remaining \textbf{20\%} for testing, utilizing five rounds of cross-validation. Consequently, for each model type, five models in total are trained on various partitions of the data. We have used the validation subset for the selection of the optimal hyperparameters. 

To meet our objectives of improving efficiency and predictive performance under constraints of limited labeled data, we reduced the labeled training samples to \textbf{20\%} of the full dataset. Thereafter, we proceeded with a progressive decrease of the size of the labeled training subset by a factor of \textbf{5\%} for each series of experiments. The lowest size of the labeled training portion used is equal to \textbf{5\%} of the full dataset. Regarding the semi-supervised training, the unused labeled training images from the training set are de-labeled and used as part of the unlabeled training set. From this unlabeled set some will be selected based on a ratio between labeled and unlabeled training samples.

Moreover, to assess the impact of increasingly scarce labeled information on the performance of the models (semi-supervised and supervised), no data augmentation techniques were employed throughout the learning phase to expand the size of the labeled training subset. All models were trained for a duration of 60 epochs and for each model type within a specific family, the five best-performing models on the validation subset were selected and kept, due to the five rounds of cross-validation. Subsequently, the evaluation of these models on the test subset provides the final average performance discussed in this study. 

The performance of the deep learning models was evaluated using two metrics namely: F1-Score and Accuracy \cite{bib67,bib68,bib69}. The first metric, defined as the harmonic mean of two metrics: \textbf{1.} Precision, which indicates the proportion of correctly classified positive predictions relative to the total positive predictions $\left(FP+TP\right)$. \textbf{2.} Recall denoting the proportion of correctly classified positive predictions relative to the real number of positives $\left(FN+TP\right)$. For the second metric it is expressed as the ratio of correctly classified predictions to the total number of predictions $\left(FN+FP+TP+TN\right)$. The equations defining these metrics are outlined as follows: 

\begin{gather} 
\textbf{\textit{F1-Score}}=2\times\left(\left(Precision\times Recall\right)/\left(Precision+Recall\right)\right) \\ \textbf{\textit{Precision}}=\left(TP\right)/\left(TP+FP\right), \textbf{\textit{Recall}}=\left(TP\right)/\left(TP+FN\right)\\ \textbf{\textit{Accuracy}}=\left(TP+TN\right)/\left(TP+TN+FP+FN\right)
\end{gather}  

\hfill \break

\subsection{Results of classification on the DeepWeeds dataset}\label{sec4.4}
\begin{figure}[H]

	\centering
	\captionsetup{justification=centering}
	\includegraphics[width=12cm,height=7cm]{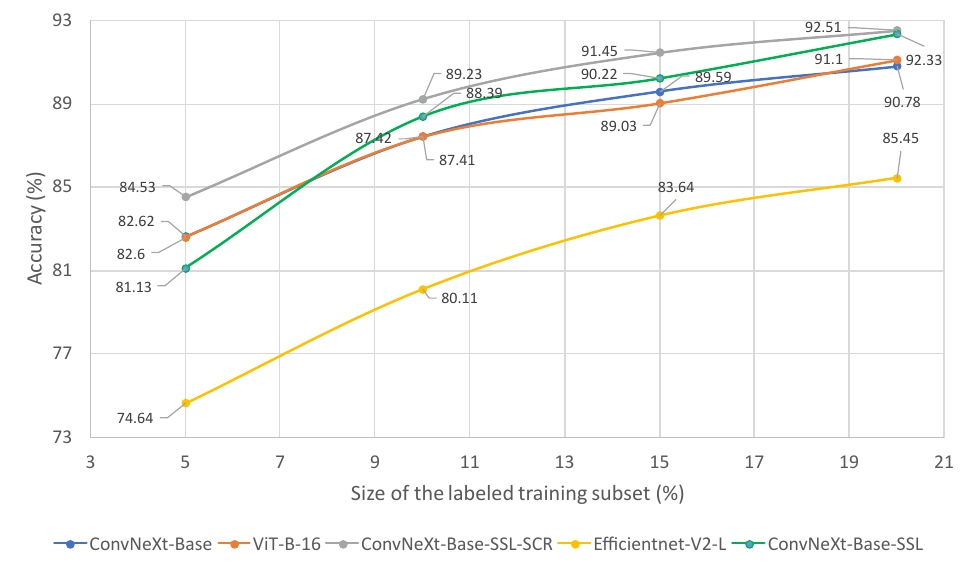}
	\caption{Average performance of the deep learning models (semi-supervised and supervised) on the test subset, based on the Accuracy metric, with different sizes of the labeled training subset. For \textbf{20\%} labeled training subset, the results of the supervised models and ConvNeXt-Base-SSL were reported from \cite{bib70} }
	\label{Figure 3}

\end{figure}

\begin{figure}[H]

	\centering
	\captionsetup{justification=centering}
	\includegraphics[width=12cm,height=7cm]{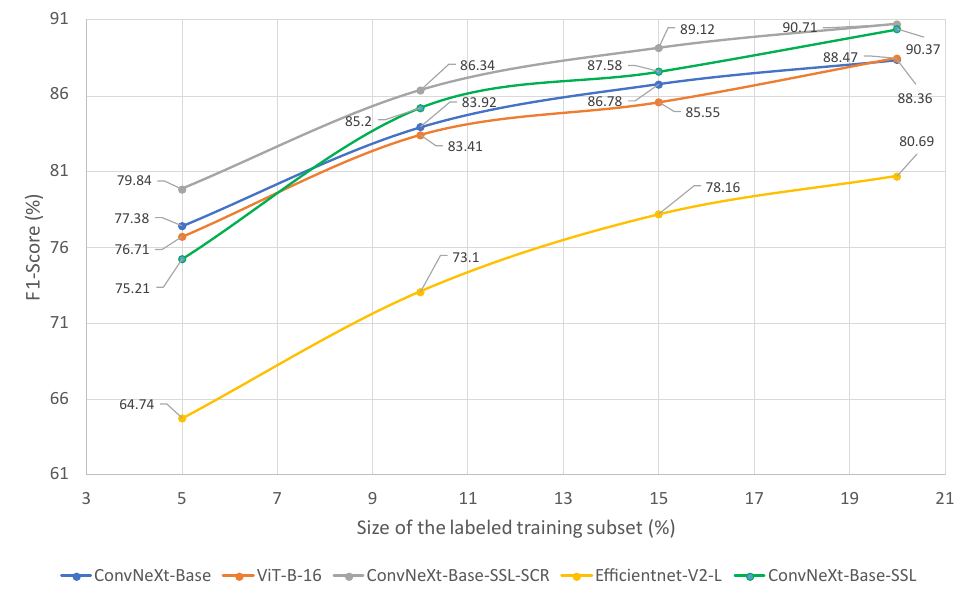}
	\caption{Average performance of the deep learning models (semi-supervised and supervised) on the test subset, based on the F1-Score metric, with different sizes of the labeled training subset}
	\label{Figure 4}

\end{figure}

Figures (\ref{Figure 3}) and (\ref{Figure 4}) depict the results of evaluations regarding the series of experiments conducted on the DeepWeeds dataset for both the semi-supervised and supervised models, based on the two metrics Accuracy and F1-Score. \textbf{ConvNeXt-Base-SSL-SCR} refers to the models trained in a semi-supervised manner using our auto-encoder architecture and the proposed learning strategy, which relies on a three terms of loss function: supervised loss, consistency regularization, and similarity loss. On the other hand, \textbf{ConvNeXt-Base-SSL} refers to the semi-supervised models trained in a similar way, through the proposed Encoder-Decoder structure incorporating two terms (supervised and consistency regularization) during the learning process, without applying the transformations of similarity.

From the obtained results, we can observe that when the size of the labeled training subset is constrained to \textbf{20\%}, the semi-supervised models (ConvNeXt-Base-SSL-SCR and ConvNeXt-Base-SSL) achieved the highest classification performances (\textbf{92.51\%} and \textbf{92.33\%}) compared with the best performing supervised models (ViT-B-16 and ConvNeXt-Base) with considerable margins in terms of both metrics. For instance, when we compare ConvNeXt-Base-SSL-SCR with Vit-B-16, we have a difference of \textbf{1.41\%} in Accuracy and \textbf{2.24\%} in F1-Score. 

Moreover, when the size of the labeled training subset becomes more constrained moving from \textbf{20\%} to \textbf{10\%}, we can observe that the semi-supervised models continue to demonstrate higher performance scores (\textbf{89.23\%} and \textbf{88.39\%}) compared to the best performing supervised models, with more increased margins regarding Accuracy and F1-Score. When the size of the labeled training subset becomes very low, i.e., \textbf{5\%}, we can see that ConvNeXt-Base-SSL underperforms in comparison to ConvNeXt-Base and ViT-B-16. In contrast, ConvNeXt-Base-SSL-SCR remains the best approach (\textbf{84.53\%}) with more important margins given the limitations on the labeled training data. Similarly, when we carry out the comparison with respect to Efficientnet-V2-L, larger profound differences of \textbf{9.89\%} in Accuracy and \textbf{15.1\%} in F1-Score.
The results of classification achieved by ConvNeXt-Base-SSL-SCR with a decreased labeled training subset of \textbf{5\%}, show the importance of having greater degree of unlabeled data utilization, to enable a more refined ability of integrating information from the unlabeled training samples.

\clearpage

\clearpage

\subsection{Noise influence with few labeled data}\label{sec4.5}

\begin{figure}[H]

	\centering
	\captionsetup{justification=centering}
	\includegraphics[width=13cm,height=3.6cm]{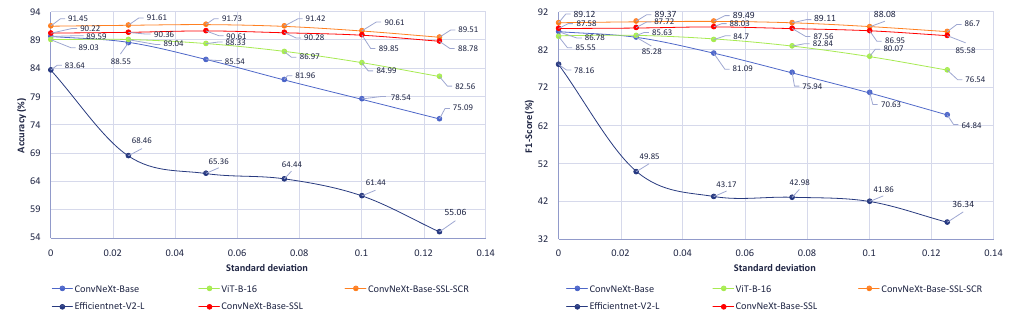}
	\caption{Average performance(Accuracy and F1-Score) of the deep learning models on the test subset, with added gaussian noise to the inputs. Data partitioning: \textbf{15\%} labeled data, \textbf{20\%} validation, \textbf{20\%} test}
	\label{Figure 5}

\end{figure}

\begin{figure}[H]

	\centering
	\captionsetup{justification=centering}
	\includegraphics[width=13cm,height=3.6cm]{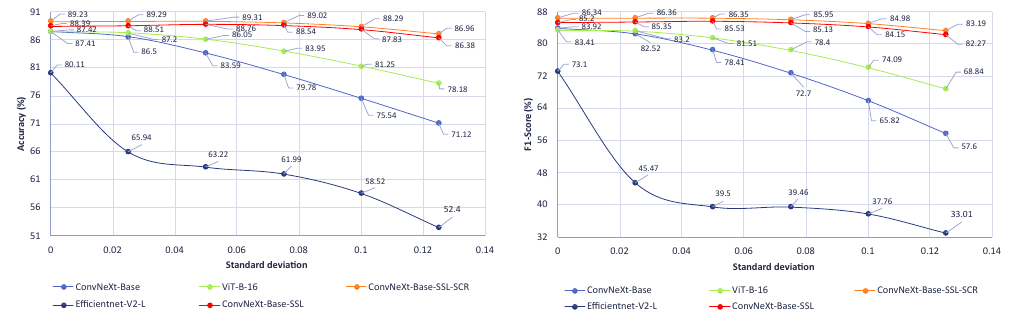}
	\caption{Average performance (Accuracy and F1-Score) of the deep learning models on the test subset, with added gaussian noise to the inputs. Data partitioning: \textbf{10\%} labeled data, \textbf{20\%} validation, \textbf{20\%} test}
	\label{Figure 6}

\end{figure}

\begin{figure}[H]

	\centering
	\captionsetup{justification=centering}
	\includegraphics[width=13cm,height=3.6cm]{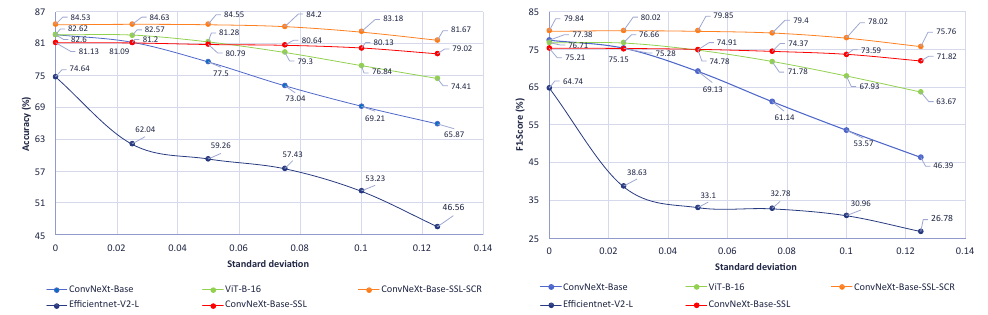}
	\caption{Average performance (Accuracy and F1-Score) of the deep learning models on the test subset, with added gaussian noise to the inputs. Data partitioning: \textbf{5\%} labeled data, \textbf{20\%} validation, \textbf{20\%} test}
	\label{Figure 7}

\end{figure}

Subsequent to the previous experiments, we have conducted an additional series of evaluations with the best-performing models saved from previous steps. The goal is to ascertain to which extent the presence of noise during inference combined with the scarcity of labeled data at training time affect the classification performance of both the semi-supervised and supervised models. These evaluations were undertaken using noisy images from the test subset, created through the addition of gaussian noise sampled from a normal distribution characterized by a fixed mean $\boldsymbol{\mu}$ and a varying standard deviation $\boldsymbol{\sigma}$.
The classification results in this context of noisy conditions, using the models trained with \textbf{15\%} and \textbf{10\%}, respectively of labeled data, are illustrated in Figures \ref{Figure 5} and \ref{Figure 6}. From these results we can observe that applying additive gaussian noise during the inference phase has a negative impact on the classification performance of purely supervised models. The impact becomes significant with raising values of the standard deviation $\boldsymbol{\sigma}$. For example, in the case where the size of the labeled training subset is limited to \textbf{15\%}, the performance of \textbf{ViT-B-16} in terms of Accuracy drops from \textbf{89.03\%} to \textbf{82.56\%}, when the standard deviation $\boldsymbol{\sigma}$ is increased to \textbf{0.125}. As for the F1-Score metric, its performance drops from \textbf{85.55\%} to \textbf{76.54\%}. In a similar manner, a smaller proportion of labeled training data leads to a further drop in performance of the supervised models. Particularly, as we can observe in Figure \ref{Figure 6}, with an increased value of $\boldsymbol{\sigma}$, the performance of ViT-B-16 drops to an Accuracy of \textbf{78.18\%} and an F1-Score of \textbf{68.84\%}.   
Figure \ref{Figure 7} depicts the results of inference in noisy conditions corresponding to the models trained using only \textbf{5\%} of labeled training samples. Similarly to the previous cases, we can identify a consistent negative correlation between the importance of noise and the drop in performance of 
the supervised models. Overall, the drop in performance of \textbf{Efficientnet-V2-L} is much more profound compared to \textbf{ViT-B-16} and \textbf{ConvNeXt-Base}, regarding the effect of higher levels of noise and the limited availability of labeled training data. Conversely, the semi-supervised models demonstrated a robust and strong classification performance relative to the increase of noise importance, which can be observed across all the different scenarios under the limitation of labeled training data. Additionally, when considering the various raising values of the standard deviation $\boldsymbol{\sigma}$, \textbf{ConvNeXt-Base-SSL-SCR} outperforms \textbf{ConvNeXt-Base-SSL} with important margins, especially in the case of less labeled training data.  

\clearpage

\subsection{Ablation studies}\label{sec4.6}

\begin{figure}[H]

	\centering
	\captionsetup{justification=centering}
	\includegraphics[width=14cm,height=5cm]{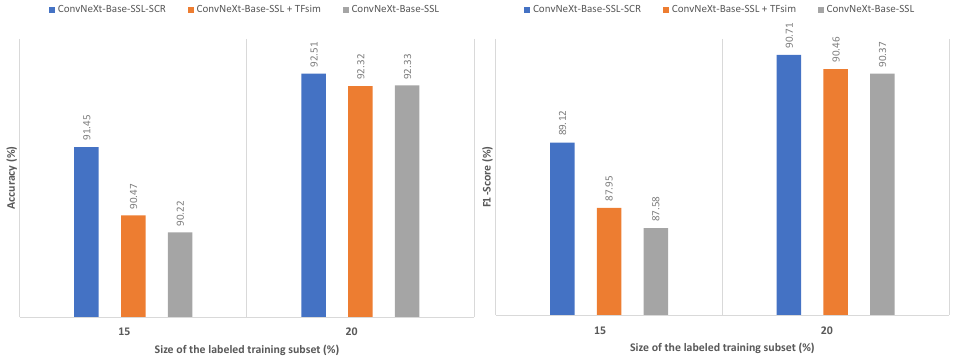}
	\caption{Comparative results of the average performance achieved by the semi-supervised models on the test subset, with and without the application of the similarity transformations (TFsim) in the loss. A ratio of one to five (\textbf{1:5}) between labeled to unlabeled data was adopted regarding both cases with respect to the size of the labeled training portion (\textbf{15\%} and \textbf{20\%})}
	\label{Figure 8}

\end{figure}

\begin{figure}[H]

	\centering
	\captionsetup{justification=centering}
	\includegraphics[width=14cm,height=5cm]{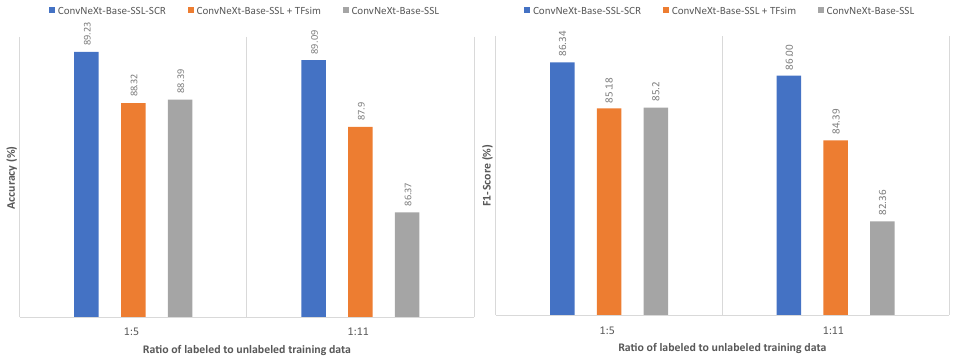}
	\caption{Comparative results of the average performance achieved by the semi-supervised models on the test subset, with and without applying the transformations of similarity in the loss with a portion of labeled training data equal to \textbf{10\%}}
	\label{Figure 9}

\end{figure}

\begin{figure}[H]

	\centering
	\captionsetup{justification=centering}
	\includegraphics[width=14cm,height=5cm]{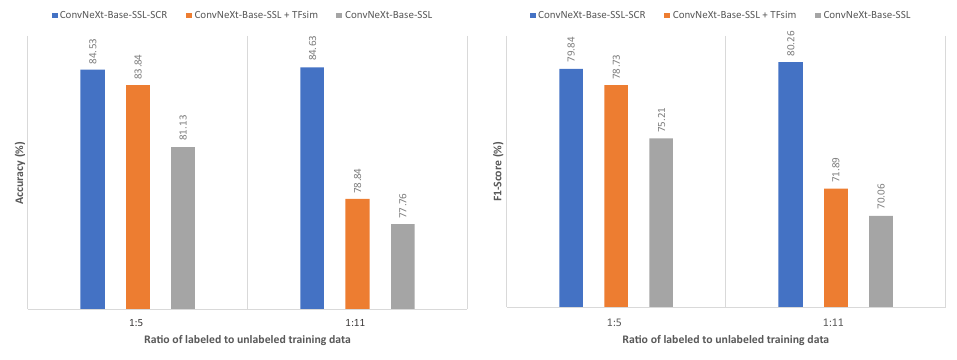}
	\caption{Comparative results of the average performance achieved by the semi-supervised models on the test subset, with and without applying the transformations of similarity in the loss with a portion of labeled training data equal to \textbf{5\%}}
	\label{Figure 10}

\end{figure}

In these experiments we realized a set of ablations focusing specifically on our semi-supervised models. Here, we have adopted the same optimal hyperparameters employed as before. This experimental setup aims to provide a more in-depth assessment of our proposed strategy, focusing on the influence of the labeled-to-unlabeled data ratio and the impact of incorporating similarity transformations into the loss function. Figure \ref{Figure 8} illustrates the results obtained by the semi-supervised models corresponding to the cases where the size of the labeled training subset is constrained to \textbf{20\%} and \textbf{15\%}, respectively, with a ratio between labeled to unlabeled data of \textbf{1:5}. When the labeled training subset is limited to \textbf{15\%}, we can observe that utilizing two terms of the total loss along with the transformations of similarity (\textbf{ConvNeXt-Base-SSL + TFsim}) yields improvement over the performance of the semi-supervised models (\textbf{ConvNeXt-Base-SSL}) trained using two terms without applying the transformations of similarity. For \textbf{20\%} of labeled data, this conclusion is less noticeable.
With respect to a limitation of \textbf{10\%} of the labeled training data, shown in Figure \ref{Figure 9}, with two different ratios utilized between labeled and unlabeled data. We can see that the impact of similarity loss depends on the ratio between labeled and unlabeled data. With a ratio of one to eleven (\textbf{1:11}), the semi-supervised models utilizing the transformations of similarity provide a better accuracy compared to the ConvNeXt-Base-SSL models with only two terms in loss. However, with a ratio of one to five (\textbf{1:5}) between the labeled and unlabeled data both variants of the semi-supervised models (ConvNeXt-Base-SSL   TFsim and ConvNeXt-Base-SSL) achieved approximately similar classification results (\textbf{88.39\%} in Accuracy and \textbf{85.20\%} in F1-Score). On the other hand, the deep semi-supervised models (ConvNeXt-Base-SSL-SCR) trained through the utilization of all three combined terms of the total loss achieved higher classification performance in comparison to the other two types of the semi-supervised models with respect to both applied ratios between labeled to unlabeled data.
Similarly, for the case where the labeled training portion is restricted to \textbf{5\%}, we can observe from Figure \ref{Figure 10}, that the ConvNeXt-Base-SSL-SCR models continue to outperform the other variants of the semi-supervised models, especially when the ratio between labeled to unlabeled data is higher. These results demonstrate the effectiveness of our proposed joint learning strategy regarding the ability to incorporate pertinent and relevant information from the unlabeled training data. 

\section{Conclusions}\label{sec5}

In this paper we introduced a deep semi-supervised approach which combines consistency regularization training with similarity learning, by leveraging our proposed auto-encoder architecture. The proposed strategy of joint learning aims at further improving the ability of incorporating information from unlabeled data, when the labeled training data become increasingly scarce. This approach was motivated by the need to accurately classify weed species from images captured in real-world conditions. To this end, we rely on deep semi-supervised models trained using both labeled and unlabeled data. The realized series of experiments on the dataset DeepWeeds showed that the models trained by integrating all three terms of our joint learning strategy provided higher classification results, particularly under conditions of limited labeled training data, outperforming both fully supervised models and semi-supervised models trained leveraging only the supervised and consistency regularization terms of the overall loss function. Furthermore, inference in noisy conditions with few labeled data demonstrated the robustness of our method compared to traditional models. Moreover, we conducted extended ablation experiments to assess the influence of the ratio between labeled to unlabeled data and the importance of the transformations applied regarding similarity learning. In future work we intend to further develop the joint learning strategy through the incorporation of pseudo labeling and extend the application of our method to object detection.

\section*{Declarations}
%
%Some journals require declarations to be submitted in a standardised format. Please check the Instructions for Authors of the journal to which you are submitting to see if you need to complete this section. If yes, your manuscript must contain the following sections under the heading `Declarations':
%
\begin{itemize}

\item \textbf{Funding}

This work was carried out as a part of DESHERBROB project funded by Region Centre-Val de Loire, France. We gratefully acknowledge its support.

\item \textbf{Conflict of interest/Competing interests}

The authors declare that they have no known competing financial interests or personal relationships that could have appeared to influence the work reported in this paper.

\item \textbf{Data availability} 

We have shared the link to the public dataset DeepWeeds used to carry out the experiments.

\item \textbf{Authors contributions}

\textbf{Farouq Benchallal}: Conceptualization, Methodology, Software, Visualization, Writing– original draft, Investigation, Validation.  
\textbf{Adel Hafiane}: Conceptualization, Methodology, Writing– review \& editing, Supervision, Project administration, Funding acquisition, Validation.  
\textbf{Nicolas Ragot}: Conceptualization, Methodology, Writing– review \& editing, Supervision, Validation.  
\textbf{Raphael Canals}: Conceptualization, Methodology, Writing– review \& editing, Supervision, Project administration, Validation. 

\end{itemize}

\bibliographystyle{unsrtnat}
\bibliography{references}  %%% Uncomment this line and comment out the ``thebibliography'' section below to use the external .bib file (using bibtex) .





% \end{thebibliography}

\end{document}